\documentclass[runningheads,a4paper]{llncs}

\usepackage{amssymb}
\usepackage{graphicx}
\usepackage{amssymb}
\usepackage{amsmath}
\usepackage{multirow}
\usepackage{caption}
\usepackage{subcaption}
\usepackage{url}

\urldef{\mailsa}\path|cr@scu.stu.edu.cn, cziyuanyang@gmail.com, zhumin@scu.edu.cn|
\urldef{\mailsb}\path|bjteoh@yonsei.ac.kr|
\newcommand{\keywords}[1]{\par\addvspace\baselineskip
\noindent\keywordname\enspace\ignorespaces#1}

\newcommand\etal{\emph{et al.}}

\begin{document}

\mainmatter  



\title{Beyond First-Order: A Multi-Scale Approach to Finger Knuckle Print Biometrics}

\titlerunning{DOTCNet}

%
%

\author{Chengrui Gao$^{1,2}$\and Ziyuan Yang$^{1}$\and Andrew Beng Jin Teoh$^{2,*}$\and Min Zhu$^{1,*}$%
}


%
\authorrunning{C. Gao et al.}


\institute{$^{1}$ College of Computer Science, Sichuan University, Chengdu 610045, China,\\
\mailsa\\
$^{2}$ School of Electrical and Electronic Engineering, Yonsei University, Seoul 03722, South Korea\\
\mailsb\\
$^{*}$ Corresponding author}


%
%

\maketitle

\vspace{-20pt}
\begin{abstract}


Recently, finger knuckle prints (FKPs) have gained attention due to their rich textural patterns, positioning them as a promising biometric for identity recognition. Prior FKP recognition methods predominantly leverage first-order feature descriptors, which capture intricate texture details but fail to account for structural information. Emerging research, however, indicates that second-order textures, which describe the curves and arcs of the textures, encompass this overlooked structural information. This paper introduces a novel FKP recognition approach, the Dual-Order Texture Competition Network (DOTCNet), designed to capture texture information in FKP images comprehensively. DOTCNet incorporates three dual-order texture competitive modules (DTCMs), each targeting textures at different scales. Each DTCM employs a learnable texture descriptor, specifically a learnable Gabor filter (LGF), to extract texture features. By leveraging LGFs, the network extracts first and second order textures to describe fine textures and structural features thoroughly.
Furthermore, an attention mechanism enhances relevant features in the first-order features, thereby highlighting significant texture details. For second-order features, a competitive mechanism emphasizes structural information while reducing noise from higher-order features. Extensive experimental results reveal that DOTCNet significantly outperforms several standard algorithms on the publicly available PolyU-FKP dataset.

\end{abstract}

\vspace{-15pt}
\section{Introduction}
\vspace{-10pt}
Biometric recognition is becoming increasingly prevalent across various application domains such as healthcare systems, public safety systems, and electronic banking~\cite{yang2024physics}.
The range of biometric technologies developed so far includes facial recognition, iris recognition, fingerprints, palmprints, and finger knuckle prints (FKP), among others~\cite{hattab2024face}.
Recently, FKP has garnered increasing research attention due to its numerous advantages~\cite{su2024complete}. For example, unlike fingerprints, which may become damaged or worn from frequent handling of objects, the surface of FKP remains largely intact. Compared to facial recognition, FKP targets are smaller and more difficult to capture maliciously, thus offering greater privacy~\cite{yang2023joint}. Additionally, the skin creases on the outer side of the FKP exhibit unique lines and wrinkles with rich textures and distinct features~\cite{fei2021jointly}. Furthermore, the data collection process for FKP is either non-contact or involves minimal contact, enhancing hygiene and making it a more user-friendly biometric modality.

In recent decades, numerous methods for FKP recognition have been developed. These approaches can be broadly categorized into two main types: (1) handcrafted methods and (2) deep learning methods. Since FKP images contain directional features similar to those in palmprints, many recent methods use palmprint encoding techniques ~\cite{guo2009palmprint}. Zhang \etal~\cite{zhang2018finger} encoded the dominant directional features of FKP images using a competitive code based on Gabor filter responses. These traditional descriptors are typically manually designed and leverage prior information. However, handcrafted algorithms often struggle to adapt to diverse modalities and varying image quality.

Deep learning methods, such as Convolutional Neural Networks (CNNs), have recently garnered significant attention in FKP recognition~\cite{li2024hand}. 
Many current CNN-based methods either utilize generic network models for training or directly adopt pre-trained models (such as VGGNet~\cite{hong2017convolutional} and ResNet~\cite{kim2018multimodal}) to extract deep features from corresponding databases. However, these generic models are primarily trained on large-scale image datasets (such as ImageNet), and the images in these datasets often exhibit a significant difference in feature distribution compared to those used in the FKP recognition task. Consequently, the performance of these models is often compromised. Thus, developing an effective neural network architecture tailored to the characteristics of FKP images is essential.
For instance, Cheng \etal~\cite{cheng2021accurate} achieved FKP recognition by investigating the learning of minimally dimensional discriminative feature vectors to represent FKP images. Li \etal~\cite{li2021learning} developed a Sparse and Discriminative Multi-modal Feature Coding (SDMFC) model for jointly learning specific and common features. Li \etal~\cite{li2021joint} proposed a Joint Discriminative Feature Learning (JDFL) model, which extracts discriminative binary codes from Gabor features for FKP recognition.
However, these methods overlook the importance of multi-order feature learning despite its crucial role in thoroughly modeling spatial correlations, thereby ultimately enhancing recognition accuracy.
As illustrated in Fig. 1, for FKP images, features processed by first-order learnable Gabor filters contain rich, detailed information. In contrast, those processed by second-order learnable Gabor filters encapsulate major structural features crucial for recognition.

To address the aforementioned issues, we introduce DOTCNet, an FKP recognition method that comprises three different scale branches, facilitating the propagation of multi-scale texture information and enhancing the features' nonlinear representation capabilities. Additionally, each branch incorporates a DTCM that employs diverse feature extraction techniques tailored to different feature orders in FKP images, thereby effectively capturing comprehensive texture and discriminative structure features. Specifically, first-order Gabor filtering is integrated with triplet attention mechanisms to enhance local and global features during the initial feature extraction phase. For second-order Gabor filtering, a competitive mechanism is utilized to select the optimal structure feature response, effectively eliminating redundant information, such as noise and irrelevant data, while preserving the important orientation of structure features.
Then, we combine first-order and second-order features, allowing higher-dimensional feature vectors to be created and offering a more comprehensive texture description.

The main contributions of this article can be summarized as follows: 
\vspace{-5pt}
\begin{itemize}
  \item We design the DTCM to utilize texture and structure information while avoiding additional noise fully. Attention and competitive mechanisms are leveraged on first-order and second-order features to focus on important characteristics.
  \item We propose an advanced FKP recognition method, DOTCNet, which combines the parallel texture-ordering feature extraction branches and the DTCM for comprehensive feature extraction.
  \item Experimental results on the open-access PolyU-FKP dataset substantiate the effectiveness of our method and demonstrate significant improvements in recognition performance.
\end{itemize}

\begin{figure}[!t]
\centering
\vspace{-5pt}
\includegraphics[height=2.5cm]{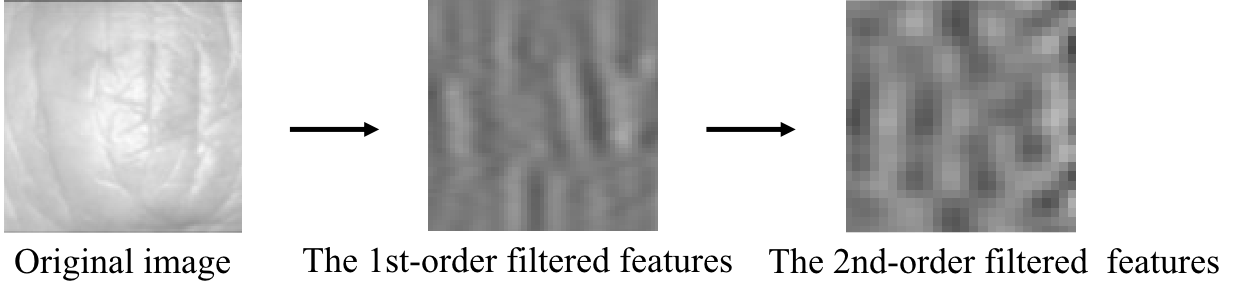}
\vspace{-5pt}
\caption{The FKP image is processed by 1st/2nd order Gabor filters}
\vspace{-20pt}
\label{fig:example}
\end{figure}

\vspace{-10pt}
\section{Proposed method}
\vspace{-5pt}
We propose DOTCNet, a network designed to enhance the global flow of multi-scale information. DOTCNet aims to achieve more refined multi-scale features and capture dual-order feature information from each scale, leading to a more comprehensive understanding of the data. The overall network architecture is shown in Fig. 2. This network consists of three branches, each containing the DTCM of different scales. For the DTCM, the first-order LGF captures features that contain rich details, while the second-order LGF captures features that contain the main structural elements. We use a spatial and channel-based attention mechanism for first-order features to extract local and global texture detail features comprehensively. Meanwhile, we focus only on the competitive relationships in the textural structure orientation of second-order features to avoid noise introduced by high-order filtering. 

\vspace{-10pt}
\subsection{Multi-scale network structure}

Different scales of DTCM are applied to the input FKP image, generating feature maps at various resolutions. This process constructs multi-scale features and integrates multi-scale contextual information. Specifically, DOTCNet is divided into three stages: Stage 1, Stage 2, and Stage 3.
Features obtained from different scales of filters — large-scale, medium-scale, and tiny-scale — contain texture information at varying scales. Tiny-scale features often have relatively large spatial extents to capture more detailed texture information, whereas large-scale features contain strong semantic information. Information at different scales is interrelated and complementary. 
Global multi-scale features are then obtained by concatenating features from different scales. These concatenated features are processed through two fully connected layers to generate the final feature vector. The following expression illustrates this process:

\vspace{-10pt}
\begin{equation}
  F = \mathrm{FC}\left( \mathrm{FC}\left( \mathrm{Concat}(F_{ls}, F_{ms}, F_{ts}) \right) \right),
\end{equation}
where FC denotes the full connection layer, $F_{ls}$, $F_{ms}$, and $F_{ts}$ represent the feature map generated by the different scale branches, Concat denotes the concatenate operation, and $F$ represents the final vector feature.

\begin{figure}[!t]
\centering
\vspace{-5pt}
\includegraphics[width=.85\textwidth]{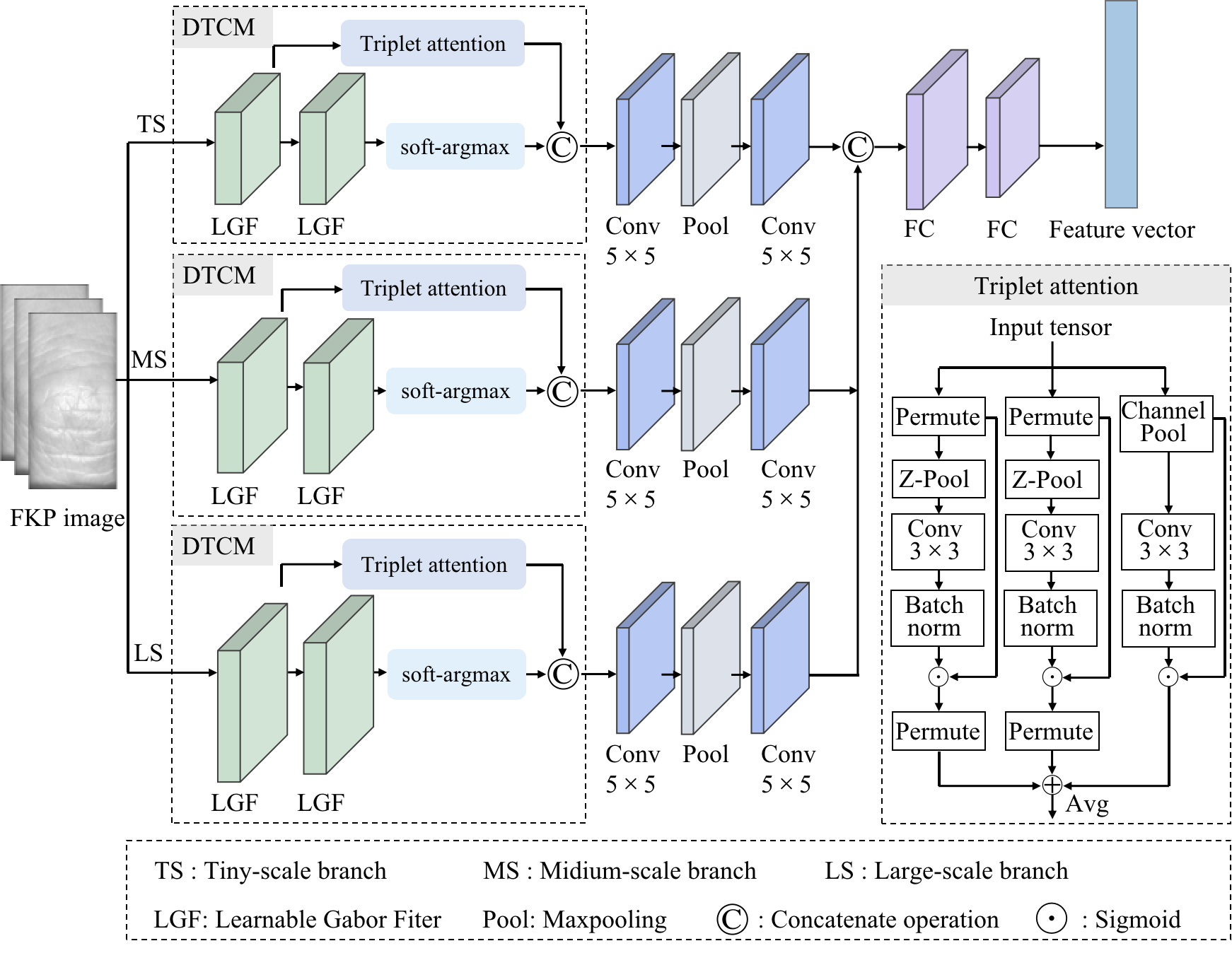}
\vspace{-5pt}
\caption{The overview of the proposed DOTCNet.}
\vspace{-25pt}
\label{fig:example}
\end{figure}

\vspace{-15pt}
\subsection{Dual-order texture competitive module}
\vspace{-3pt}

For each DTCM, the process includes capturing the initial detail texture features through first-order LGF and the initial structure features through second-order LGF. Then, the attention and competition mechanisms are used for the two order features, respectively. Taking the large-scale branch as an example, dual-order features are generated by cascading first-order features and second-order features, and the expression is as follows:
\vspace{-3pt}
\begin{equation}
  F_{ls} = \mathrm{Concat}(F_{1}, F_{2}),
\end{equation}
where $F_{ls}$ denotes the dual-order features in large-scale branch, $F_{1}$ represents the first-order detailed texture features, $F_{2}$ signifies the second-order structure features.

The Gabor filter feature extractor utilized in this paper is extensively employed in image processing because of its biological relevance and robust texture extraction capabilities. The Gabor filter is mathematically defined as follows:

\vspace{-10pt}
\begin{equation}
g(x, y; \lambda, \theta, \phi, \sigma, \gamma) = \exp \left( -\frac{x'^2 + \gamma^2 y'^2}{2\sigma^2} \right) \exp \left( i \left( 2\pi \frac{x'}{\lambda} + \phi \right) \right),
\end{equation}
where $x' = x \cos \theta + y \sin \theta$ and $y' = -x \sin \theta + y \cos \theta$. (x, y) denotes the pixel position. $\lambda$ denotes the wavelength of the sinusoidal plane wave component of the Gabor function. $\theta$ specifies the angle of the plane wave, while $\phi$ indicates the phase shift. Additionally, $\gamma$ denotes the ellipticity of the Gaussian support of the Gabor function, and $\sigma$ determines the standard deviation of the Gaussian filter within the Gabor function.

An effective Gabor filter requires an appropriate combination of parameters to match the given task. Previous studies~\cite{aliradi2024novel} have typically used handcrafted methods for parameter selection, relying on inherent rules to set the parameters manually and cannot guarantee effectiveness for the current task. To overcome this limitation, we employ the LGF as the texture extractor~\cite{chen2019lgcn}, which facilitates learning optimal parameters ($\lambda, \sigma, \gamma$) to extract texture features. In this paper, we employ the real part of the Gabor function, using Gabor filters with sizes of 7, 17, and 35 for tiny-scale, medium-scale, and large-scale textures. The filters are set to 12, 36, and 6, respectively.

\vspace{-15pt}
\subsubsection{The first-order detailed texture feature extraction}

We first use first-order LGF to extract the edge and texture information of the image preliminarily. The expression is as follows:

\vspace{-10pt}
\begin{equation}
\chi = \mathrm{LGF}(X),
\end{equation}
where X represents the input FKP image, $\chi $ is the initial detail texture features through first-order LGF, and LGF($\cdot$) is the LGF operation.

For the generated feature $\chi $, we establish dimensional dependencies through rotation operations and residual transformations, capture cross-dimensional interactions to calculate attention weights, and encode channel and spatial information through the Triplet attention \cite{misra2021rotate}. The architecture of the triplet attention module is shown in Fig. 2.
Given an input feature $\chi \in \mathbb{R}^{C \times H \times W}$, we obtain the $(C, H)$-branching identity $\hat{\chi}_1$ by rotating it $90^\circ$ anti-clockwise along the $H$ axis. We rotate $\chi$ $90^\circ$ anti-clockwise along the $W$ axis to obtain the $(C, W)$ branch feature map $\hat{\chi}_2$. $\chi$ is the $(H, W)$ branch feature, denoted as $\chi_3$. Then, the Z-Pool process connects each branch's average pooling and maximum pooling results, which can be expressed as:
\vspace{-5pt}
\begin{equation}
\hat{\chi}_i^{*} = \mathrm{Z\_Pool}(\hat{\chi}_i) = [\mathrm{MaxPool}(\hat{\chi}_i), \mathrm{AvgPool}(\hat{\chi}_i)],
\end{equation}
where $\hat{\chi}_i^{*}$ means the feature map through the Z-Pool process.

Finally, the interaction of spatial and channel attention across different dimensions can be represented as:

\begin{equation}
F_{1} = \frac{1}{3} (\overline{\hat{\chi}_1 \sigma (\psi_1 (\hat{\chi}_1^{*}))} + \overline{\hat{\chi}_2 \sigma (\psi_2 (\hat{\chi}_2^{*}))} + \chi \sigma (\psi_3 (\chi^{*}))),
\end{equation}
where $F_{1}$ represents the first-order detailed texture feature, $\sigma$ represents the sigmoid activation function. $\psi_1$, $\psi_2$, and $\psi_3$ represent the standard two-dimensional convolutional layers defined by kernel size 3 in the three branches of triplet attention.
The overline represents rotating the input tensor $90^\circ$ clockwise.

\vspace{-10pt}
\subsubsection{The second-order structure feature extraction}

After the process above, the first-order LGF captures detailed features and ensures a balance between local and global information through the attention mechanism. Building on this, the second-order LGF further strengthens important structural features, making the final feature representation more discriminative. The specific expression is as follows:

\vspace{-10pt}
\begin{equation}
g = \mathrm{LGF}(F_{1}),
\end{equation}
where $F_{1}$ represents the generated first-order texture feature, $g$ means the initial structure features through second-order LGF.

To capture the important structure information, soft competitive code (SCC) \cite{liang2021compnet} is introduced to extract the ordering relationship as the feature using the Softmax function. The process is formulated as follows:

\begin{equation}
F_{2} = \mathrm{softmax} (g)
\end{equation}
where $F_{1}$ is the input of the competitive mechanism, respectively. Softmax($\cdot$) denotes the competition extraction process along the channel dimension.

\vspace{-10pt}
\section{Experiments and Results}
\vspace{-5pt}
\subsection{Datasets and Experimental Settings}

This section introduces the PolyU Finger Knuckle Print Database (PolyU-FKP) used for experimental analysis~\cite{zhang2009finger}. The database comprises FKP images collected by the Hong Kong Polytechnic University using a low-resolution camera in a peg-free environment. The dataset involves 148 individuals and includes images of the left index FKP, left middle FKP, right index FKP, and right middle FKP. The images were captured in a contactless mode with a resolution of 110 × 220 pixels in BMP format. For each FKP image, there are 12 images, resulting in a total of 591 finger classes. We randomly selected six samples from each FKP to create the FKP training set while reserving the remaining samples for the FKP testing phase.

Our method is implemented using the PyTorch framework and optimized with the Adam optimizer~\cite{kingma2014adam}, utilizing a learning rate of 0.01 and a batch size 1024. The experiments are conducted on an NVIDIA GTX 3090 GPU. In this paper, we develop a loss function by integrating cross-entropy loss with contrastive loss~\cite{gao2024scale}. For comparison, other deep learning methods are implemented by replacing the network while keeping all other parameters consistent.

\vspace{-10pt}
\begin{table}[!t]
\centering
\vspace{-15pt}
\caption{Comparison of the proposed method with state-of-the-art methods.}
\label{branch}
\setlength{\tabcolsep}{6pt} 
\begin{tabular}{lccccc}
\hline
\multirow{2}{*}{Methods} & \multicolumn{5}{c}{EER(\%)}   \\ \cline{2-6} 
 & Left index & Left middle & Right index & \begin{tabular}[c]{@{}l@{}}Right middle\end{tabular} & \multicolumn{1}{c}{Total} \\ \hline
BOCV~\cite{guo2009palmprint}                     & 8.753          & 8.712           & 8.845           & 8.667                                                      & 8.657                             \\
ResNet18~\cite{kim2018multimodal}         & 5.349          & 4.392           & 5.545           & 4.657                                                      & 3.484                             \\
DenseNet101~\cite{song2019finger}              & 5.686          & 5.968           & 5.799           & 6.708                                                      & 4.725                             \\
VGG16~\cite{hong2017convolutional}                    & 6.550          & 4.842           & 6.250           & 4.649                                                      & 4.039                             \\
Compnet~\cite{liang2021compnet}                  & 2.558          & 3.228           & 4.940           & 3.498                                                      & 2.934                             \\
PCANet-FKP~\cite{attia2022deep}     & 4.542          & 2.614           & 4.054           & 3.307                                                      & 2.431                             \\
CO3Net~\cite{yang2023co}                  & 4.748          & 4.354           & 5.405           & 5.291                                                      & 4.294                             \\
CCNet~\cite{yang2023comprehensive}                    & 2.477          & 3.059           & 3.941           & 3.042                                                      & 2.622                             \\
Ours   & \textbf{2.327}  & \textbf{2.571}  & \textbf{2.909 }   &\textbf{ 2.664}                                                      & \textbf{2.186}                             \\ \hline
\end{tabular}
\vspace{-5pt}
\end{table}

\begin{figure}[t]
    \centering
    \vspace{-5pt}
    \begin{subfigure}[t]{0.3\textwidth}
        \centering
        \includegraphics[width=\textwidth]{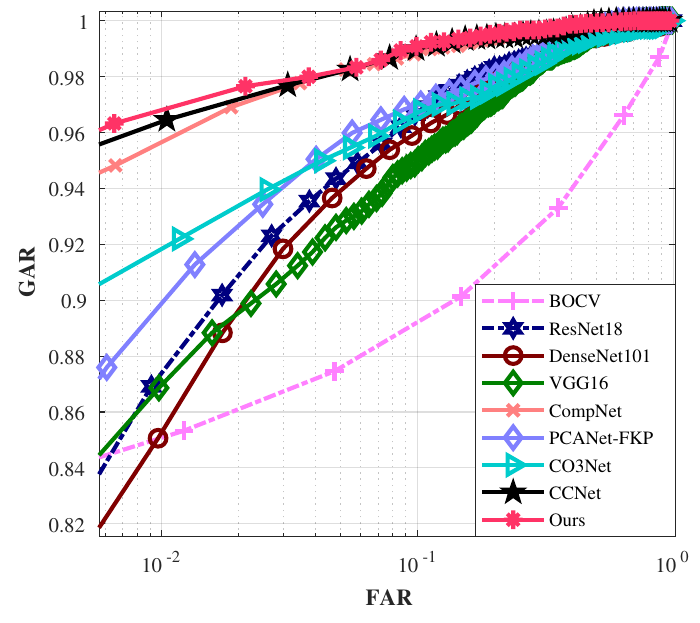}
        \caption{Left index FKP}
        \label{fig:polyU}
    \end{subfigure}
    \hfill
    \begin{subfigure}[t]{0.3\textwidth}
        \centering
        \includegraphics[width=\textwidth]{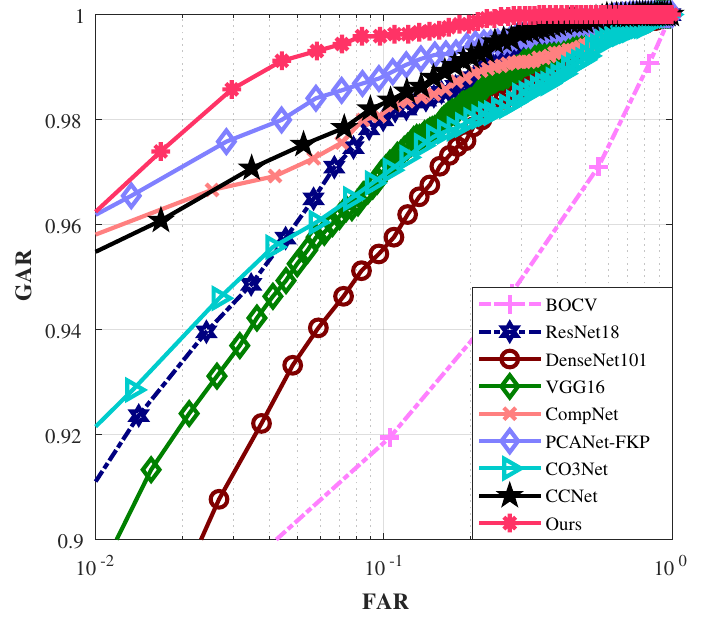}
        \caption{Left middle FKP}
        \label{fig:iitd}
    \end{subfigure}
    \hfill
    \begin{subfigure}[t]{0.3\textwidth}
        \centering
        \includegraphics[width=\textwidth]{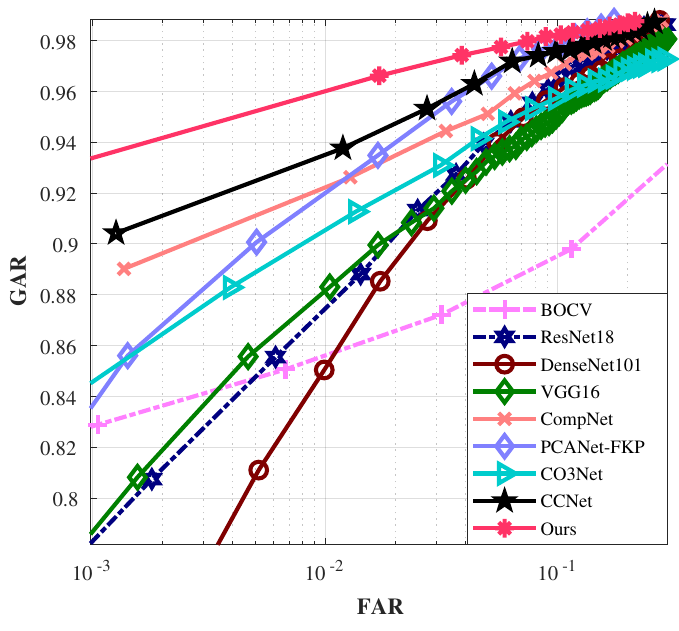}
        \caption{Right index FKP}
        \label{fig:nir}
    \end{subfigure}
    
    \begin{subfigure}[t]{0.3\textwidth}
        \centering
        \includegraphics[width=\textwidth]{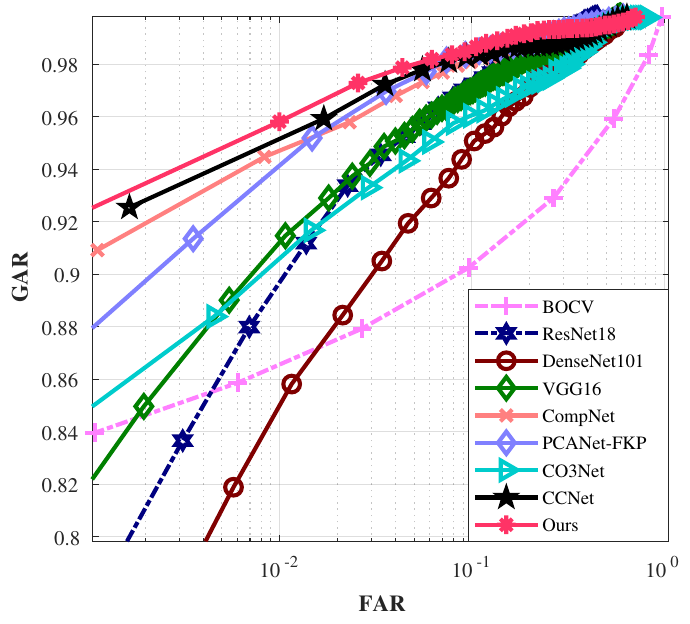}
        \caption{Right middle FKP}
        \label{fig:red}
    \end{subfigure}
    \begin{subfigure}[t]{0.3\textwidth}
        \centering
        \includegraphics[width=\textwidth]{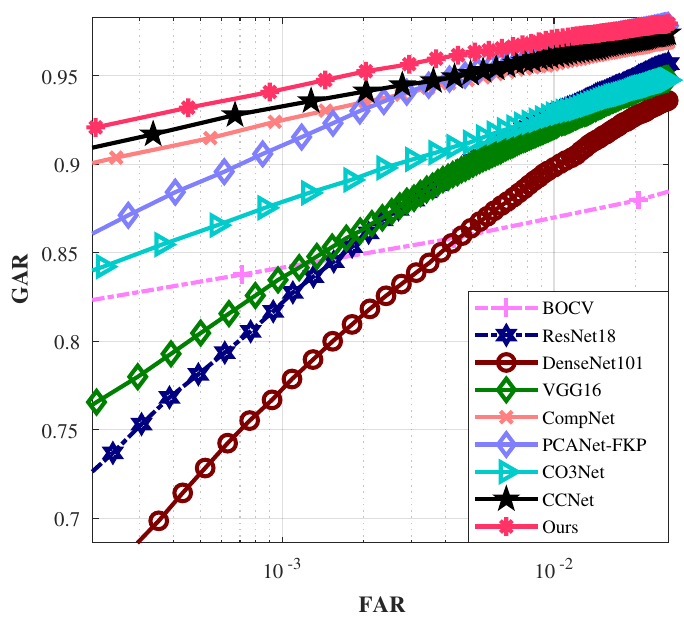}
        \caption{All dataset}
        \label{fig:green}
    \end{subfigure}
\vspace{-5pt}
    \caption{The ROC curves of the proposed and compared methods on PolyU-FKP.}
    \vspace{-5pt}
    \label{fig:roc}
\end{figure}

\vspace{-5pt}
\subsection{Recognition Performance}
\vspace{-2pt}
We compared the proposed method with the classical FKP recognition methods BOCV~\cite{guo2009palmprint}, ResNet18~\cite{kim2018multimodal}, DenseNet101~\cite{song2019finger}, VGG16~\cite{hong2017convolutional}, Compnet~\cite{liang2021compnet}, and PCANet-FKP~\cite{attia2022deep}. Moreover, we also implemented two advanced texture description methods, CO3Net~\cite{yang2023co} and CCNet~\cite{yang2023comprehensive}. Through experiments conducted on the PolyU-FKP dataset (comprising left index FKP, left middle FKP, right index FKP, and right middle FKP), we evaluated the performance of these methods. Table 2 presents the FKP recognition performance of various models on the dataset. For instance, our method achieved an Equal Error Rate (EER) of 2.186 on the overall FKP dataset, which is significantly lower than the second-ranked PCANet-FKP, representing an improvement of 11.2\%. Our method also demonstrated superior EER performance across the PolyU-FKP dataset subset. Overall, the proposed method achieved the lowest EER across all test datasets, indicating superior accuracy and robustness in FKP recognition tasks.

To further validate the effectiveness of our method, we plotted the corresponding Receiver Operating Characteristic (ROC) curves, as shown in Fig. 3. The closer the curve is to the top left corner of the plot, the better the performance of the corresponding algorithm. These curves illustrate that our method exhibits superior performance across different thresholds, significantly outperforming other comparative methods. Compared with traditional techniques and other deep learning-based approaches, our method achieves the lowest EER values and the most optimal ROC curves on the PolyU-FKP dataset, showcasing its robustness and efficacy.

\begin{table}[!t]
\centering
\vspace{-22pt}
\caption{Accuracies (\%) and EERs (\%) obtained with different scale branches on PolyU-FKP.}
\setlength{\tabcolsep}{8pt}
\begin{tabular}{ccccc}
\hline
Large-scale               & Midium-scale              & Tiny-scale                & ACC(\%)            & EER(\%)            \\ \hline
\checkmark & $\times$                  & $\times$                  & 98.84          & 4.308          \\
$\times$                  & \checkmark & $\times$                  & 99.40          & 2.949          \\
$\times$                  & $\times$                  & \checkmark & 99.32          & 2.980          \\
\checkmark & \checkmark & $\times$                  & 99.37          & 2.702          \\
$\times$                  & \checkmark & \checkmark & 99.51              & 2.338              \\
\checkmark & \checkmark & \checkmark & \textbf{99.60} & \textbf{2.186} \\ \hline
\end{tabular}
\end{table}

\vspace{-5pt}
\begin{table}[!t]
\centering
\vspace{-15pt}
\caption{Accuracies (\%) and EERs (\%) obtained with different mechanisms on PolyU-FKP.}
\label{branch}
\setlength{\tabcolsep}{12pt} 
\begin{tabular}{cccccc}
\hline
\multicolumn{2}{c}{1st-order feature} & \multicolumn{2}{c}{2nd-order feature} & \multirow{2}{*}{ACC(\%)} & \multirow{2}{*}{EER(\%)} \\ \cline{1-4}
TAM    & CM    & TAM    & CM    &     &         \\ \hline
\checkmark     & \checkmark    & $\times$     & $\times$ & 99.37        & 2.881 \\
\checkmark     & \checkmark    & \checkmark     & \checkmark    & 99.49        & 2.610                \\
\checkmark     &    $\times$   & \checkmark     &   $\times$   & 99.51   & 2.570                \\
  $\times$  & \checkmark    &   $\times$  & \checkmark    & 99.49        & 2.585    \\
   $\times$   & \checkmark    & \checkmark     &    $\times$   & 99.54  & 2.468     \\
\checkmark     &    $\times$  &   $\times$  & \checkmark    & \textbf{99.60 }   & \textbf{2.186}      \\ \hline
\vspace{-20pt}
\end{tabular}
\vspace{-5pt}
\end{table}

\vspace{-10pt}
\subsection{Ablation Experiments}
\vspace{-5pt}
\subsubsection{The efficiency of multi-scale branches}
To validate the necessity of multi-scale branches in recognition performance, we conducted several ablation experiments on the PolyU-FKP dataset. The entire experimental process ensured consistent network layers to ensure result comparability. The experimental results are shown in Table 1. First, we tested the recognition performance using only the large, medium, and tiny individual branches separately. The experimental results showed that the medium-scale branch outperformed the large-scale and tiny-scale branches, performing the best. Next, based on the medium-scale branch, we added the large-scale and tiny-scale branches for combination testing. The results indicated that the combination of medium-scale and tiny-scale branches outperformed the combination of medium-scale and large-scale branches, further validating the importance of the tiny-scale branch in the multi-scale structure. Based on the performance of the three branches in the experiments and the analysis of their importance, we allocated appropriate network layers to each branch. Specifically, the large-scale branch was assigned six layers, the medium-scale branch 36 layers, and the tiny-scale branch 12 layers. This allocation of network layers effectively improved the model's recognition performance.

\vspace{-15pt}
\subsubsection{The efficiency of dual-order texture extraction}

We conduct ablation experiments to evaluate the importance of dual-order texture features and the contributions of the Triplet Attention Mechanism (TAM) and the Competition Mechanism (CM) for first-order and second-order textures. These experiments isolate and elucidate the impact of each mechanism on overall model performance. The configurations of the ablation experiments on the PolyU-FKP dataset are as follows: (1) Only retaining the first-order texture extraction, (2) TAM and CM for both first-order and second-order textures, (3) TAM for both first-order and second-order textures, (4) CM for both first-order and second-order textures, (5) CM for first-order texture and TAM for second-order texture, and (6) TAM for first-order texture and CM for second-order texture.

As shown in Table 2, configuration (1) is compared with (2), and it is concluded that retaining dual-order features is significantly better than retaining only first-order features. The highest performance was observed in configuration (5), indicating that TAM effectively captured the essential features of first-order textures. In contrast, the CM was more suitable for handling the complexity and feature differentiation in second-order textures. This suggests that first-order textures benefit more from TAM’s focused attention on salient features, whereas second-order textures require effectively differentiating between significantly varying directional structure features.

\vspace{-5pt}
\section{Conclusion}
\vspace{-5pt}
In this paper, we propose a novel FKP recognition network termed DOTCNet. This network incorporates multi-scale branches and the DTCM to achieve comprehensive feature extraction.
For DTCM, detailed texture features are captured by the first-order LGF. At the same time, inter-dimensional dependencies are established through rotational operations and residual transformations within the triplet attention mechanism, preserving the cross-dimensional texture features. Subsequently, structural features are captured by the second-order LGF and processed via a competitive mechanism to differentiate feature directions.
Finally, the dual-order features are concatenated to achieve comprehensive feature extraction.
To validate the effectiveness of the proposed method, we conducted extensive experiments on the open-access dataset. The experimental results indicate that our method has significant advantages over several other methods.
Future research directions include establishing the consistency of templates for left and right FKP images. This approach utilizes registered left (right) FKP images to verify the identity of right (left) FKP images.


\end{document}